%% file: main.tex
\documentclass{impfai-report}
\usepackage{mathtools}
\usepackage{booktabs,tabularx,array,ragged2e,longtable}
\hypersetup{pdftitle={TCMScience: An Autonomous Scientist for Traditional Chinese Medicine},pdfauthor={Yanlan Kang; Ruiqi Liu; Xinpei Gu; Tzu-Yi Lei; Shuai Xu; Xukun Zhang; William Cheng-Chung Chu},pdfsubject={IMPF-AI Technical Report: evidence-governed scientific execution}}
\reportshorttitle{TCMScience}
\newcolumntype{Y}{>{\RaggedRight\arraybackslash}X}
\newcolumntype{P}[1]{>{\RaggedRight\arraybackslash}p{#1}}
\newcommand{\code}[1]{\texttt{#1}}
\newcommand{\Ind}{\mathbb{I}}
\input{metrics}

\title{TCMScience: An Autonomous Scientist for Traditional Chinese Medicine}
\author{%
  Yanlan Kang\textsuperscript{1,\textdagger} \and
  Ruiqi Liu\textsuperscript{2,\textdagger} \and
  Xinpei Gu\textsuperscript{3,\textdagger} \and
  Tzu-Yi Lei\textsuperscript{4} \\
  Shuai Xu\textsuperscript{5,*} \and
  Xukun Zhang\textsuperscript{6,*} \and
  William Cheng-Chung Chu\textsuperscript{7,*}}
\affiliation{1}{Institute of Medical Philosophy \& Future AI (IMPF-AI)}
\affiliation{2}{Shanghai Medical College, Fudan University}
\affiliation{3}{Guangdong Provincial Key Laboratory of New Drug Screening and Guangdong--Hong Kong--Macao Joint Laboratory for New Drug Screening, School of Pharmaceutical Sciences, Southern Medical University}
\affiliation{4}{Department of Computer Science, Tunghai University, Taichung, Taiwan}
\affiliation{5}{Shanghai Ziranerran Traditional Chinese Medicine Foundation}
\affiliation{6}{Li Ka Shing Faculty of Medicine, The University of Hong Kong}
\affiliation{7}{Fuyao University of Science and Technology}
\authornote{\textsuperscript{\textdagger}Equal contribution.\quad
  \textsuperscript{*}Co-corresponding authors.}
\correspondence{Shuai Xu, \href{mailto:nanfeng4115@163.com}{nanfeng4115@163.com};\\
  Xukun Zhang, \href{mailto:xukunzhg@hku.hk}{xukunzhg@hku.hk};
  William Cheng-Chung Chu, \href{mailto:20240831@fyust.edu.cn}{20240831@fyust.edu.cn}.}
\date{2026-10-15}

\begin{document}
\maketitle
\begin{abstract}
Traditional Chinese medicine (TCM) research connects classical texts, medicinal entities,
molecular assays, computational predictions, and clinical studies. An autonomous research
agent must preserve the inferential boundaries between these evidence types while coordinating
analysis across heterogeneous resources. We introduce TCMScience, a scientific knowledge-discovery
agent that separates language-model proposals from execution and release authority. Its
architecture combines a trusted kernel, a scientific compiler, and claim-support contracts:
plans are compiled into typed operations, execution is bounded by explicit permissions,
and candidate findings are checked against source provenance, study design, and claim scope
before release. The full capability registry contains \NConnectors{} public-source connectors
with \NOperations{} typed operations across \NConnectorDomains{} domains and \NTools{} native
tools across \NToolDomains{} domains; the TCM data hub catalogs \NDatabases{} databases,
\NDatasets{} dataset specifications, and \NFormulaRecords{} formula-source records. These
resources support model-independent workflows through browser and local execution interfaces.
In a network-pharmacology analysis of Gegen Qinlian Tang, replacing the whole-proteome
background with the assayed-protein background changed the number of significant pathways
from 66 to zero; incorporating inactive screening records identified nine pathways,
including drug-metabolism signals retained as mechanistic hypotheses. A governance ablation
over \NMutants{} constructed erroneous outputs spanning \NErrorClasses{} classes released
121 errors without governance, 57 with the domain layer, seven with the kernel, and zero
with both layers. All \NControls{} compliant controls were retained. These results establish
complementary control of known failure modes and show how explicit evidence contracts change
the conclusions an agent is permitted to report.

\end{abstract}
{\footnotesize
\par\noindent\textbf{Studio:} \url{https://science.impf.ai}\quad
\textbf{Code:} \url{https://github.com/IMPF-AI-Institute/TCMScience}
}
\clearpage

\section{Introduction}
Scientific knowledge discovery in traditional Chinese medicine (TCM) requires reasoning
across sources that answer different questions. A classical passage establishes what a text
records; a chemical-composition resource associates a constituent with a medicinal source;
a binding assay measures an interaction under a specified protocol; a molecular model
predicts a structural or functional property; and a clinical study estimates an outcome in
a defined population. Combining these records does not erase the conditions under which
each was produced. The central challenge is therefore to compose useful research workflows
without converting evidence from one domain into a stronger claim in another.

Existing TCM artificial-intelligence research provides essential components of this
workflow. Machine-learning systems analyze diagnostic modalities\cite{t}; knowledge graphs
organize herbs, formulas, syndromes, diseases, and their relations\cite{q}; and domain
language models, including ZhongJingGPT and HuatuoGPT, support medical dialogue and reasoning
\cite{zhongjinggpt,huatuogpt}. Graph-based retrieval systems such as OpenTCM connect these
structured resources to natural-language interaction\cite{h}. These advances make relevant
knowledge more accessible, while leaving the validity of a composed scientific inference
as a distinct problem.

Scientific agents extend language models from answering questions to acting on them.
Coscientist and ChemCrow combine planning with chemical tools\cite{coscientist,chemcrow};
The AI Scientist and Co-Scientist develop research and hypothesis-generation workflows
\cite{aiscientist,coscientist2}; and Biomni coordinates biomedical tools, data, and executable
analysis\cite{b}. As these capabilities expand, a technically successful tool call and a
scientifically supported conclusion must remain separate objects. Correct execution of
a docking program does not establish a therapeutic effect. Accurate retrieval of a paper
does not authorize extending its result to an unstudied population.

Network pharmacology makes this distinction particularly consequential. It links formula
constituents, targets, and pathways to generate mechanistic hypotheses\cite{hopkins,li-zhang},
but enrichment depends on the sampled proteins, annotation density, inactive records, and
multiple-testing procedure\cite{timmons,wijesooriya,bh}. A research agent can produce a fluent
mechanism narrative even when the apparent signal is attributable to the assay background.
The failure occurs at the boundary between computation and interpretation, where additional
retrieval alone cannot supply the missing inferential justification.

\begin{figure}[tbp]
\centering
\includegraphics[width=\textwidth]{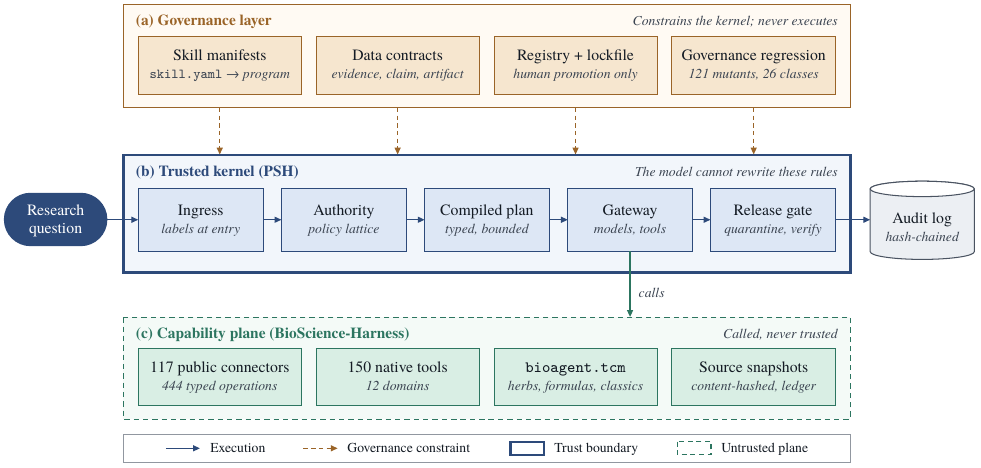}
\caption{TCMScience framework. The governance layer constrains the trusted kernel; the
capability plane is reached through governed gateways. Solid arrows show execution and
dashed arrows show constraints. Counts use the full-registry convention in
Table~\ref{tab:inventory}. Governance evaluation refers to the constructed regression corpus.}
\label{fig:framework}
\end{figure}

TCMScience addresses this boundary through a governed execution architecture
(Figure~\ref{fig:framework}). The language model proposes a scientific program; a
compiler checks its types, effects, and evidence requirements; a trusted kernel authorizes
execution; and the release path verifies the resulting claims. Domain knowledge enters
through typed entities and contracts, including medicinal identity, processing state,
formula composition, exposure, and source independence. These checks make TCM-specific
scientific errors inspectable and actionable within the same runtime that executes the
analysis.

The report makes three contributions. First, it specifies the separation of generative
reasoning, computational authority, and scientific release as explicit architectural
interfaces. Second, it formalizes evidence support using typed claim obligations and
provenance-bearing inference paths, distinguishing serial dependencies from alternative
support and contextual citations. Third, it demonstrates the consequences of these controls
through a source-backed network-pharmacology analysis and a layer-wise and gate-wise
regression ablation. The emphasis is on the scientific decisions changed by the architecture.

\section{Related Work and Technical Positioning}
\subsection{Domain models, knowledge graphs, and retrieval}
TCM diagnostic models focus on prediction from clinical or sensory inputs\cite{t}.
Knowledge-graph systems address entity integration and relational access\cite{q}, while
TCM-focused language models and GraphRAG systems improve domain interaction
\cite{zhongjinggpt,huatuogpt,h}. TCM-QA further shows the importance of evaluating knowledge
across question formats and languages\cite{l}. TCMScience uses these capabilities within
a different control abstraction: a retrieved passage, a model-generated interpretation,
and a releasable claim have separate types and validation states. The runtime preserves
that distinction through execution and bilingual presentation.

\subsection{Scientific agents and executable research}
Tool-using scientific agents demonstrate that language models can coordinate chemistry,
biomedical analysis, and hypothesis generation\cite{coscientist,chemcrow,aiscientist,coscientist2,b}.
TCMScience treats the model and connected tools as replaceable components behind governed
interfaces. Its contribution is the integration of authorization, scientific type checking,
source-state verification, and domain-specific claim validation. This positioning is
architectural: the experiments below isolate governance effects on a fixed corpus and
analysis choices in a fixed case study, rather than rank foundation models or general-purpose
agents.

\subsection{Provenance and valid biological interpretation}
Provenance supports inspection of how data and analyses produce an artifact, consistent
with FAIR principles\cite{fair}. Resources such as LOTUS, UniProt, STRING, Reactome, and
Open Targets provide complementary biological representations\cite{lotus,uniprot,string,reactome,opentargets}.
Their relations have different meanings: occurrence, identifier equivalence, protein
association, pathway membership, and target--disease association are not interchangeable.
TCMScience therefore preserves relation semantics alongside source identity. Statistical
controls operate on this typed representation so that a change in background population
or evidence definition remains visible in the resulting conclusion.

\section{System Architecture}
\subsection{Components and interfaces}
\begin{figure}[p]
\centering
\includegraphics[width=\textwidth]{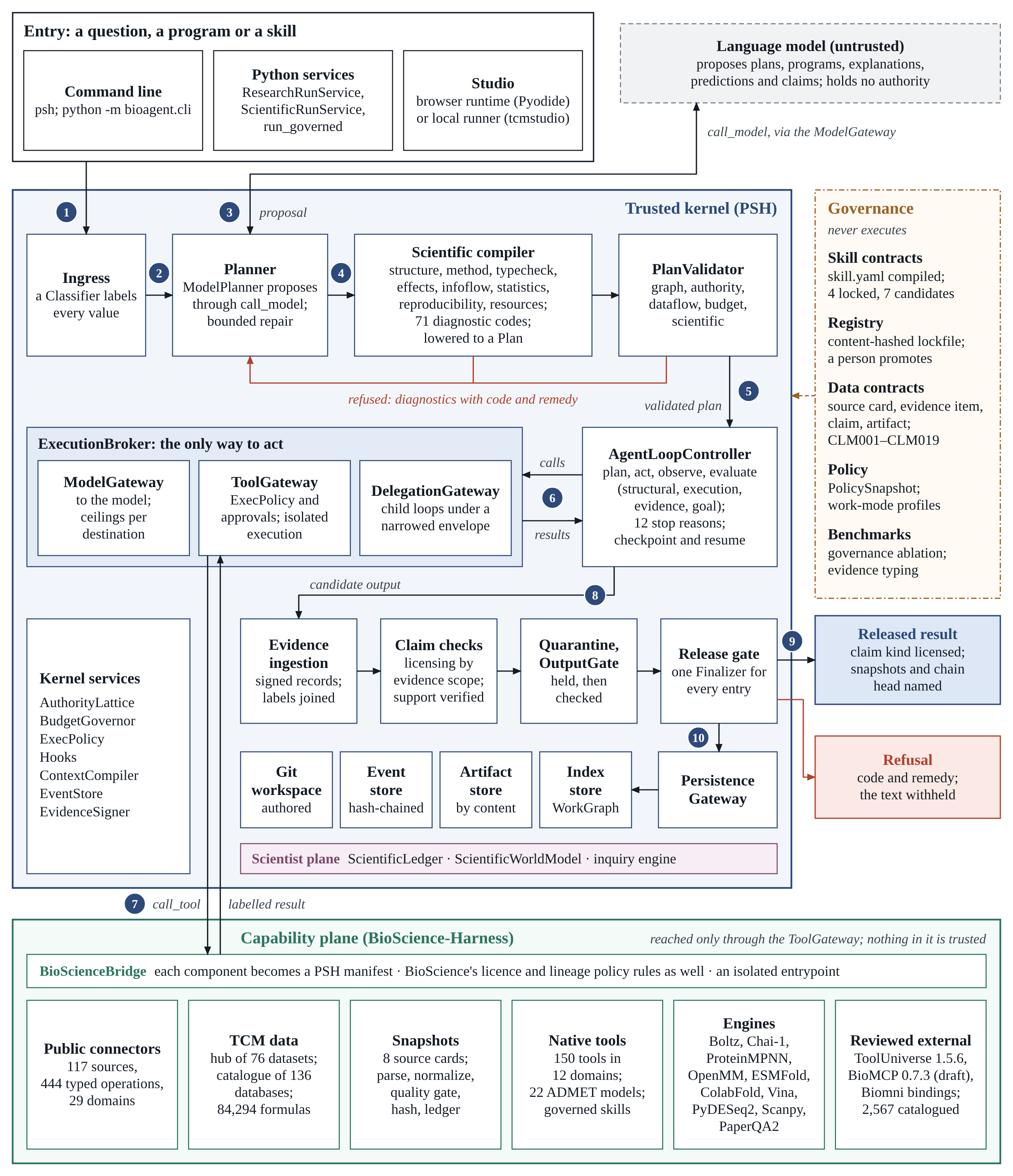}
\caption{Detailed execution architecture. Proposals are compiled and validated before the
broker can act; evidence, claim, and output checks precede release. The governance panel
separates locked skills from candidate compositions. Capability-plane counts are registry
inventories; source cards identify the snapshot subsystem, and external entries retain their
own readiness state. The runtime admits capabilities under the active policy and dependencies.}
\label{fig:architecture}
\end{figure}

Figure~\ref{fig:architecture} details the components and data flow.
TCMScience comprises three coordinated planes. The \emph{scientist plane} maintains the
research question, hypotheses, protocols, predictions, observations, and deviations. The
\emph{capability plane}, implemented in BioScience-Harness, supplies connectors, native
tools, parsers, datasets, and domain workflows. The \emph{governance plane} maintains policies,
skill manifests, version locks, contracts, and validators. The Physician-Scientist Harness
(PSH-Harness) supplies the trusted kernel through which governed execution and release pass.

TCMScience Studio provides the interactive interface; Python and command-line entry points
provide programmatic access to the same research components. The implementation namespace
\code{bioagent} belongs to BioScience-Harness, \code{psh} to PSH-Harness, and
\code{tcmstudio} to the Studio local Runner. These names identify software components within
one architecture. Interactive exploration, scripted analysis, and batch snapshot construction
share the evidence and provenance model while exposing the controls appropriate to each entry
point.

\subsection{Authority confinement and execution mediation}
The kernel labels incoming data, freezes the applicable policy, and binds the research task
to an execution envelope. For an operation $o$, let $A_{\rm host}$, $A_{\rm project}$,
$A_{\rm run}$, and $A_{\rm skill}$ denote the allowed actions at the corresponding scopes.
Its effective authority is
\begin{equation}
A_{\rm eff}=A_{\rm host}\cap A_{\rm project}\cap A_{\rm run}\cap A_{\rm skill},
\qquad \mathrm{Authorize}(o)=\Ind[o\in A_{\rm eff}].
\label{eq:authority}
\end{equation}
A skill whose declared requirements exceed this intersection is rejected during validation.
Delegation inherits an equal or narrower envelope, and consumption is charged to the applicable
budget scopes. Neither a generated plan nor retrieved text can extend the authority in
Equation~\ref{eq:authority}.

An execution broker mediates model, tool, and delegated calls. It records requested actions,
bindings, outcomes, and failure states. The model gateway controls inference calls, the tool
gateway controls computation and external access, and the delegation gateway constrains child
workflows. Candidate outputs remain quarantined until validation. Audit events form an
append-only hash chain, and released artifacts record the chain head used by the run.
This provides tamper-evident lineage within the trusted host and policy boundary; operating-system
isolation is a separate deployment property, not an implication of the word ``kernel''.

\subsection{Scientific compilation}
The scientific compiler transforms a proposed program into a typed, bounded execution plan.
The intermediate representation distinguishes evidence, claims, operations, bindings, and
effects. Evidence carries study design, subject, population, intervention, comparator,
outcome, and provenance refinements. Claims specify the corresponding scope and the kind of
conclusion requested. Compilation checks graph structure, type compatibility, permissions,
data flow, statistical obligations, reproducibility metadata, and resource limits before the
program reaches execution.

The compiler makes two separations explicit. Human-readable skill descriptions explain a
method, while machine-readable manifests define its executable contract. Similarly, successful
computation produces a candidate artifact, while release requires a separate validation
verdict. Diagnostics identify the violated obligation and the corresponding remedy, such as
resolving a medicinal entity, supplying an admissible source, or narrowing a claim. A repaired
plan is revalidated under the same authority ceiling.

\subsection{Scientific state and governed skills}
The scientist plane stores hypotheses and protocols as versioned records, associates
observations with the hypotheses they address, and records deviations from planned analyses.
The inquiry controller uses predictions specified before execution to select informative
analyses. Its epistemic state remains separate from claim authorization: a high posterior
belief does not change the study design required to support a conclusion.

Four core TCM skill contracts encode recurring domain obligations: entity normalization,
evidence retrieval, network-pharmacology analysis, and safety-evidence assessment. They are
composite workflows over the capability registry, rather than a count of all available
analytical operations. Entity normalization exposes unresolved alternatives; retrieval
preserves evidence design and empty results; network pharmacology separates measured activity
from prediction; and safety assessment distinguishes recorded risk, unmeasured outcomes,
and absent records. Skill promotion and version locking determine which compositions enter
the governed release path.

\FloatBarrier
\section{Evidence Semantics and Claim-Support Contracts}
\subsection{Evidence kinds and claim scope}
A scientific claim is represented by its content, claim kind, subject, population,
intervention or exposure, comparator, outcome, direction, and asserted certainty. An evidence
record identifies the source and locator, source version, study design, measured variables,
and verification state. A relation retains both \emph{how the assertion was generated}
(for example, measurement, curation, prediction, or text co-occurrence) and \emph{the study
design behind it}. These fields prevent a curated prediction from being relabeled as an
experiment merely because it appears in a database.

Evidence kinds form a compatibility relation with claim kinds, rather than a single ordered
ladder. A randomized trial can support a bounded human-outcome claim, but does not establish
what an ancient text said. Conversely, a classical passage can directly support textual
attribution while supplying no estimate of human efficacy. Table~\ref{tab:claims} summarizes
the principal obligations. Source verification establishes identity and integrity; it does
not by itself establish adequate study quality or inferential support.

\begin{table}[htbp]
\centering\small
\caption{Representative claim-support obligations. Permissions are conditional on source
verification and matching scope; evidence design alone is not sufficient for release.}
\label{tab:claims}
\begin{tabularx}{\textwidth}{P{0.19\textwidth}P{0.27\textwidth}Y}
\toprule
\rowcolor{impflight}
\textbf{Claim kind} & \textbf{Relevant evidence} & \textbf{Required boundary}\\
\midrule
Textual attribution & Located classical passage or source record & Preserve wording,
edition, locator, and attribution; do not infer a measured treatment effect.\\\addlinespace
Traditional use & Documented historical or practice source & Retain the traditional indication,
formula identity, preparation, and source context.\\\addlinespace
Mechanistic hypothesis & Docking, target prediction, network analysis, or exploratory
preclinical evidence & Label the conclusion as a hypothesis and preserve computational or
experimental conditions.\\\addlinespace
Experimental mechanism & Relevant in-vitro or animal experiment & Match biological system,
intervention, concentration, outcome, and causal design; do not transfer directly to patients.\\\addlinespace
Human association or effect & Appropriately designed human study & Match population,
intervention, comparator, endpoint, and study design; retain uncertainty and bias assessment.\\\addlinespace
Safety finding & Explicitly measured and reported safety outcome & Name the event, observation
window, and population; an unmeasured outcome remains unknown.\\
\bottomrule
\end{tabularx}
\end{table}

\subsection{Formal release specification}
Let $c$ be a candidate scientific claim, $r$ its run, and $C(c)$ the finite set of cited
records. Citations have explicit roles: supporting, contextual, or opposing. Let
$\mathcal{S}_{\pi}(c)$ be the family of candidate supporting subsets drawn from $C(c)$ under
policy $\pi$; every member is nonempty and contains only records offered as support.
Define $V(e)$ as successful verification of record identity, integrity, and source binding;
$\mathrm{Suff}_{\pi}(S,c)$ as satisfaction of the complete claim-support contract by a
supporting set $S$; and $G_{\pi}(r,c,C)$ as successful execution, artifact, scope, and
whole-citation-set checks. A precise release condition is
\begin{equation}
\begin{aligned}
\mathrm{Release}_{\pi}(c,r)={}&\Ind[C(c)\ne\varnothing]
\left(\prod_{e\in C(c)}V(e)\right)\,G_{\pi}(r,c,C(c))\\[-2pt]
&{}\times\Ind\!\left[\exists S\in\mathcal{S}_{\pi}(c):
\mathrm{Suff}_{\pi}(S,c)=1\right].
\end{aligned}
\label{eq:release}
\end{equation}
All factors in Equation~\ref{eq:release} are Boolean. The explicit nonempty-citation condition and the requirement for
a nonempty sufficient supporting set make an evidence-free claim fail even though an empty
product equals one. If no sufficient set exists, the existential condition is false.
The gate also requires the released wording to match the validated claim and records
unresolved conflicts and incomplete required assessments as failures of the applicable
contract. A cited opposing result cannot be ignored by selecting a favorable subset.

The supporting set can contain one directly sufficient study or several complementary
records evaluated under a declared synthesis procedure. Sufficiency requires claim-kind
compatibility, matched scope, valid inferential dependencies, and the quality and independence
conditions specified by the contract. It is not defined by taking the highest evidence grade
or by counting citations. In particular, a verified randomized trial is not automatically
sufficient for an unrestricted efficacy statement.

\subsection{Serial dependencies, parallel support, and context}
The release contract distinguishes two forms of evidence composition. In a serial derivation
$p$, every indispensable edge must be valid and must preserve its allowed claim type. In
a set of alternative derivations $\mathcal{P}_{\pi}(c)$, any complete sufficient derivation
can meet the support obligation:
\begin{equation}
\begin{aligned}
\mathrm{PathValid}_{\pi}(p)&=\bigwedge_{a\in p}\mathrm{Admissible}_{\pi}(a),\\
\mathrm{Supported}_{\pi}(c)&=
\bigvee_{p\in\mathcal{P}_{\pi}(c)}
\left[\mathrm{PathValid}_{\pi}(p)\wedge\mathrm{Sufficient}_{\pi}(p,c)\right].
\end{aligned}
\label{eq:paths}
\end{equation}
The disjunction over no derivations in Equation~\ref{eq:paths} is false. A branch requiring several complementary
records uses conjunctive obligations inside that branch; alternatives are disjunctive only
when each can independently satisfy the declared inference. The ``weakest link'' principle
therefore applies to indispensable dependencies, not to every item in a bibliography.

For example, a properly scoped clinical-effect claim can be supported by an admissible trial
and accompanied by a docking study used only to discuss a mechanistic hypothesis. The docking
study is verified and assessed against its own contextual role; it neither licenses the
clinical claim nor invalidates the trial's support merely by being cited. If the claim instead
asserts that the predicted interaction caused the clinical effect, the interaction becomes
an indispensable causal dependency and must satisfy that stronger obligation. Context cannot
serve as a route around the support contract.

\subsection{Domain-specific overclaiming controls}
TCM evidence requires constraints beyond generic citation matching. Medicinal names retain
botanical identity, plant part, processing state, and aliases. Formula-level statements require
formula-level support: a result for one constituent cannot be transferred to a decoction by
name association alone. Processing changes are explicit, so evidence for a processed medicinal
is not silently reused for its raw counterpart. Experimental concentration is retained when
interpreting exposure, and repeated access to the same study through multiple databases does
not create independent replication.

The output layer checks the natural-language statement as well as the structured claim.
It detects scope drift in subjects, populations, outcomes, direction, and certainty; checks
citation identity and quoted passages; and rejects assertions about outcomes that the cited
study did not measure. Thus, the absence of an adverse-event entry is represented as missing
evidence, while an observed negative assay remains a measured result under its assay conditions.
This distinction is enforced in the artifact, rather than delegated to cautionary wording
in the final response.

\section{Capability Registry and Reproducible Data Integration}
\subsection{A single inventory convention}
Table~\ref{tab:inventory} defines the full-registry inventory used throughout this report.
The public-source registry catalogs \NConnectors{} connectors and \NOperations{} typed
operations across \NConnectorDomains{} domains. Native computation contributes \NTools{}
tools across \NToolDomains{} domains. The TCM hub catalogs \NDatabases{} database resources,
\NDatasets{} dataset specifications, and \NFormulaRecords{} formula-source records. These
are different counting units with overlapping coverage; they are not additive counts of
independently validated datasets. A formula record retains its source and variant identity,
so the formula-record count does not assert that every entry is a distinct canonical formula.

\begin{table}[htbp]
\centering\small
\caption{Unified full-registry inventory. Counts describe registered resources and operations.
Runtime admission is evaluated separately for each requested capability.}
\label{tab:inventory}
\begin{tabularx}{\textwidth}{P{0.23\textwidth}P{0.18\textwidth}Y}
\toprule
\rowcolor{impflight}
\textbf{Registry object} & \textbf{Count} & \textbf{Counting unit}\\
\midrule
Public-source connectors & \NConnectors & Registered source connectors in the general capability plane.\\\addlinespace
Typed operations & \NOperations{} / \NConnectorDomains{} domains & Declared source operations with typed inputs and outputs.\\\addlinespace
Native analytical tools & \NTools{} / \NToolDomains{} domains & Registered computational tool implementations.\\\addlinespace
TCM database catalog & \NDatabases & Source resources with access and provenance metadata.\\\addlinespace
TCM dataset specifications & \NDatasets & Dataset acquisition or integration specifications.\\\addlinespace
Formula-source records & \NFormulaRecords & Source-retaining formula entries, including variants.\\\addlinespace
Core snapshot source cards & \NSnapshotCards & Cards in the network-pharmacology snapshot subsystem.\\\addlinespace
Core TCM skill contracts & \NCoreSkills & Entity, retrieval, network-pharmacology, and safety workflows.\\
\bottomrule
\end{tabularx}
\end{table}

Registration, execution readiness, and scientific validation are separate states. A call is
admitted only when its operation, source access, required dependencies, and policy conditions
are satisfied. This separation allows a large discovery catalog to coexist with narrowly
scoped execution contracts. Catalog growth cannot silently authorize additional tools or
change the evidence available to an ongoing analysis.

\subsection{Snapshot scope and source roles}
The eight core snapshot source cards cover NPASS, CMAUP, LOTUS, BindingDB, PubChem BioAssay,
STRING, Reactome, and Open Targets. They identify the source-specific contracts of the
network-pharmacology subsystem, rather than the total source inventory. UniProt's reviewed
human accessions provide an auxiliary identity reference for protein mapping; this reference
is versioned with the inputs without being counted as a ninth card in that subsystem.
Table~\ref{tab:sources} relates these resources to the scientific operations they support.

\begin{table}[htbp]
\centering\small
\caption{Source roles and preserved relation semantics in the network-pharmacology workflow.}
\label{tab:sources}
\begin{tabularx}{\textwidth}{P{0.23\textwidth}P{0.29\textwidth}Y}
\toprule
\rowcolor{impflight}
\textbf{Resource} & \textbf{Data role} & \textbf{Preserved interpretation}\\
\midrule
NPASS, CMAUP, LOTUS & Natural-product occurrence, composition, and available activity records &
Separate medicinal occurrence, constituent identity, and assay-derived activity.\\\addlinespace
BindingDB, PubChem BioAssay & Binding measurements or screening outcomes & Retain assay IDs,
units, qualifiers, target identity, active/inactive status, and assay conditions.\\\addlinespace
UniProt reference mapping & Reviewed human protein identity & Map to unique primary accessions;
retain unresolved or excluded mappings in quality-control counts.\\\addlinespace
STRING & Protein association network & Preserve association confidence; an association edge
does not establish direct binding or causal direction.\\\addlinespace
Reactome & Protein-to-pathway membership & Retain pathway version, annotation provenance, and
the declared enrichment background.\\\addlinespace
Open Targets & Target--disease associations & Keep genetic, literature, and other evidence
channels distinct, with channel-specific thresholds.\\
\bottomrule
\end{tabularx}
\end{table}

Access modes include live API queries, downloaded snapshots, authorized manual imports, and
cataloged resources with restricted or unavailable access. The analytic workflow parses raw
records, normalizes typed nodes and relations, applies quality checks, computes content hashes,
and registers the resulting snapshot in a ledger. Claims refer to record-level identifiers
and relations within that source state. Snapshot loading rechecks content against the manifest
and ledger; a modified source state requires a new snapshot identity.

Source stewardship is part of the same contract. Cards retain attribution, access conditions,
and reuse permissions. A source's research availability and redistribution permission remain
separate fields. The snapshot pipeline records how an authorized source was obtained, while
the release artifact retains the lineage needed to inspect and reproduce the derived result.

\subsection{Identity resolution and reproducibility artifacts}
Identifier alignment is an analytic operation. Compound records retain source identifiers
and standardized chemical cross-references; protein mapping uses reviewed primary accessions;
and formula relations preserve source variants and medicinal roles. During development,
auditing the PubChem gene-to-UniProt crosswalk revealed mappings to secondary or unreviewed
accessions that did not align with the pathway reference. Restricting mappings to unique
reviewed human primary accessions restored consistent cross-source joins. The case study in
Section~\ref{sec:case} uses the corrected mapping.

A reproducible run binds the scientific program and skill version to source snapshots,
parameters, random seed, software identity, and output digests. The network-pharmacology
pipeline writes constituent, target, enrichment, and network tables together with structured
claims, release decisions, and a provenance record. The provenance record includes snapshot
identifiers, source allowance, purpose, and ledger head. These artifacts separate the observed
data, the analysis applied to them, and the claims permitted by that analysis.

\Needspace{6\baselineskip}
\section{Execution Lifecycle and Bilingual Research Interface}
\subsection{A common execution and release lifecycle}
\begin{figure}[!htbp]
\centering
\includegraphics[width=\textwidth]{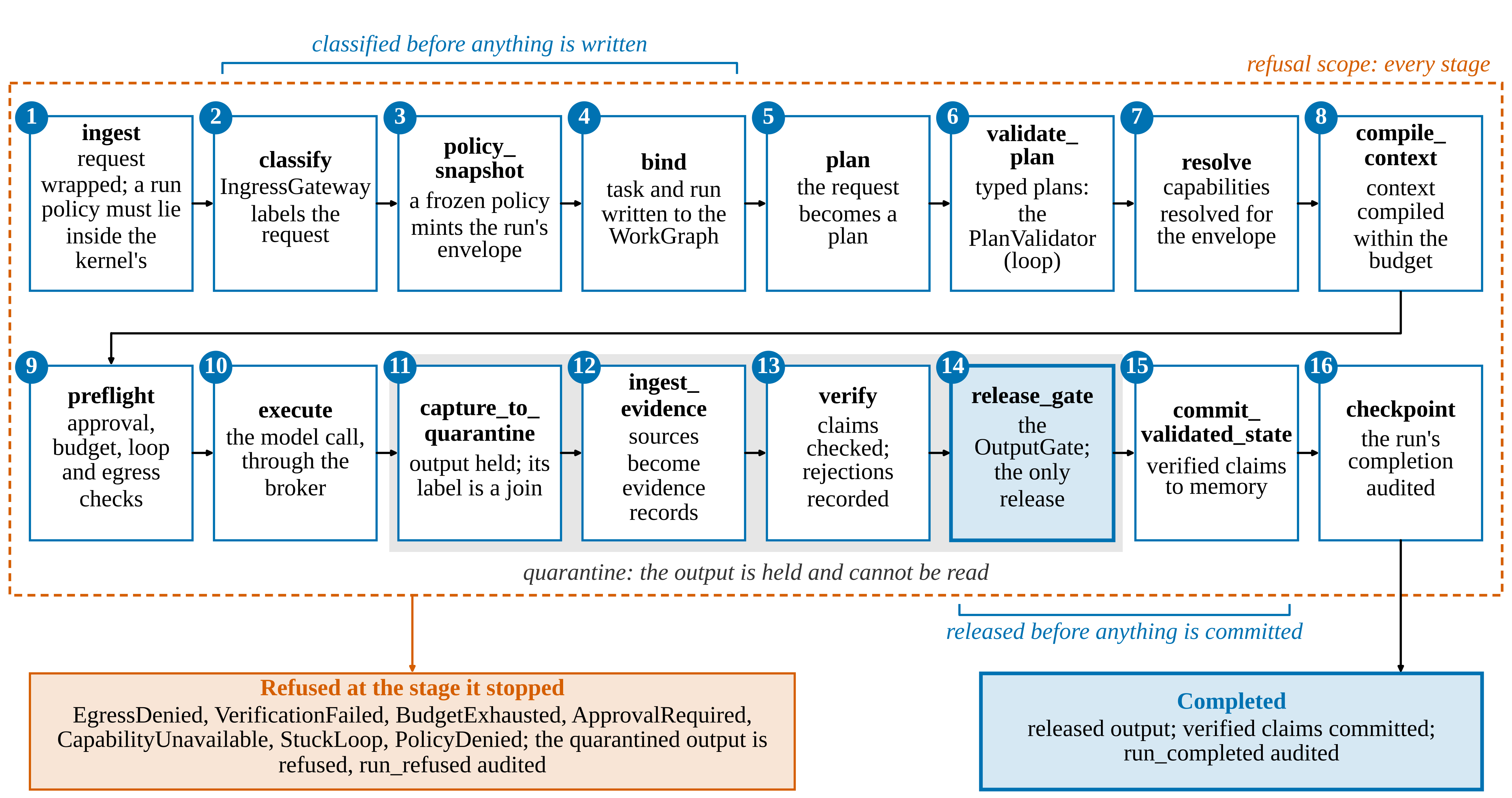}
\caption{The 16-stage governed lifecycle. Numbered stages and arrows indicate execution order. The
release gate is the boundary between candidate output and a released artifact; validated
state is committed only after that boundary. Failed obligations produce a recorded refusal
at the relevant stage. Revisions re-enter plan validation without expanding authority.}
\label{fig:lifecycle}
\end{figure}

Figure~\ref{fig:lifecycle} presents the 16-stage lifecycle, from request ingestion to an
audited completion checkpoint. Policy binding precedes planning; plan validation and capability
resolution precede execution; and evidence ingestion, verification, and release precede
commitment of validated scientific state. The same finalization boundary applies to a
single tool workflow and to a multi-step research loop.

Validation failures return typed diagnostics at the stage where they occur. An unmet
permission, unavailable capability, exhausted budget, or invalid evidence path prevents
release under the active contract. A revised plan returns to validation, and subsequent
iterations retain the authority ceiling and audit lineage. Quarantine distinguishes
intermediate generated content from material authorized for scientific reporting.

\subsection{Independent model, compute, and network controls}
Studio separates three configuration dimensions: the model endpoint used for reasoning,
the backend used for computation, and the hosts permitted for external access. The default
inference service is exposed under the deployment alias Tao-S1. Compatible external or local
model endpoints can replace it without changing the scientific contracts. Model identity
and routing belong to run configuration, while claim authorization remains a kernel function.

Lightweight Python computation runs in a browser Web Worker through Pyodide/WebAssembly.
The local Runner executes native Python workloads and integrates installed computational
engines for larger analyses. Model calls and source queries retain their own network
permissions even when computation is local. This makes execution location and information
flow explicit architectural properties rather than assumptions inferred from the user
interface. Scientific capability is determined by the selected backend, installed
dependencies, and granted authority.

\subsection{Bilingual identity preservation}
Chinese and English interaction share canonical entities, source locators, units, and claim
types. Translation changes explanatory text while retaining the identifiers that determine
which evidence supports a statement. Ambiguous names such as \zh{姜} (jiang) require a distinction
between \zh{生姜} (fresh ginger) and \zh{干姜} (dried ginger), and formula names retain their textual
source or canonical record identifier. Processing and near-name distinctions remain visible
in both languages.

Table~\ref{tab:bilingual} shows representative input-to-contract mappings. These are examples
of interface semantics, not additional empirical cases. The contract is independent of
whether the question requests a Chinese or English explanation. Safety-oriented retrieval
can preserve the classical ``Eighteen Incompatible Medicinals'' (Shibafan, \zh{十八反}) relation
as a traditional source statement while keeping measured adverse outcomes separate.

\begin{table}[!htbp]
\centering\footnotesize
\caption{Bilingual requests mapped to language-independent research obligations.}
\label{tab:bilingual}
\begin{tabularx}{\textwidth}{P{0.36\textwidth}Y}
\toprule
\rowcolor{impflight}
\textbf{Representative Chinese request} & \textbf{Executable obligation and English interpretation}\\
\midrule
\zh{区分葛根芩连汤成分的实测靶点与预测靶点。} & Separate measured targets from predicted
targets; preserve source records, study design, and relation type.\\\addlinespace
\zh{请确认}“\zh{姜}”\zh{对应的中药实体。} & Resolve ``jiang'' to a canonical medicinal entity;
request disambiguation when fresh and dried ginger remain possible.\\\addlinespace
\zh{解释通路富集的背景集与阴性记录。} & Identify the enrichment background, retain inactive
assay outcomes, and state the multiple-testing procedure.\\\addlinespace
\zh{用英文总结，但保留原始名称和出处。} & Translate the explanation while preserving original
entity names, source locators, units, and validated claim scope.\\
\bottomrule
\end{tabularx}
\end{table}

\Needspace{6\baselineskip}
\section{Case Study: Evidence-Constrained Network Pharmacology}
\label{sec:case}
\subsection{Formula, records, and analytic specification}
We analyzed Gegen Qinlian Tang (\zh{葛根芩连汤}), represented as the four-herb formula comprising
Puerariae Lobatae Radix (\zh{葛根}), Scutellariae Radix (\zh{黄芩}), Coptidis Rhizoma (\zh{黄连}), and
Glycyrrhizae Radix et Rhizoma (\zh{甘草}). Composition records from NPASS, CMAUP, and LOTUS yielded
1,792 formula-associated constituents. These are source-linked constituents, not measured
concentrations in an administered preparation. Measured potency records covered 391 human
proteins, of which 237 had qualifying activity values at or below $10\,\mu\mathrm{M}$.
The hit definition used recorded IC$_{50}$, K$_i$, K$_d$, or EC$_{50}$ values; a censored value
that did not establish the threshold was not treated as a qualifying hit.

Pathway analysis used Reactome membership, retained the declared background population,
and applied Benjamini--Hochberg (BH) correction across all tested pathways\cite{bh}.
Pathways were selected using their original annotation size of 5--500 proteins; tests used
members falling within the specified background. An annotation-degree-matched null further
controlled the tendency of extensively annotated proteins to contribute to many pathways.
These controls address distinct selection mechanisms and are retained in the analysis
specification. Table~\ref{tab:case} summarizes the contrasts.

\subsection{Assay coverage changes the pathway conclusion}
With the full human Reactome annotation as background (12,155 proteins), 66 of 1,684 tested
pathways passed the specified BH and degree-matched-null criteria. Restricting the background
to the 391 proteins actually assayed changed the test family to 959 pathways, none of which
passed the same decision procedure. The broad-background result was therefore sensitive to
assay coverage and did not support a background-robust pathway claim.

The assay-matched result also identifies a concrete information constraint: 237 of 391
assayed proteins were recorded as hits, a hit fraction of 60.6\%. Such a positive-enriched
collection provides limited contrast for overrepresentation. A nonsignificant result under
this background does not establish biological absence. The governed default retains the
analysis and its provenance while withholding a positive pathway-mechanism claim. This is
a scientific consequence of the analysis specification, rather than a failure to produce
a result.

\subsection{Inactive records recover a distinct screening signal}
The PubChem BioAssay analysis retained 143,679 explicitly classified results across 1,029
human proteins: 10,696 active and 132,983 inactive records. Under the default coverage
threshold of at least 20 tested constituents per protein, 472 proteins contributed 51,304
compound--protein tests, including 3,357 active results (6.54\%). Analysis at the test level
preserves information that would be lost by labeling a protein positive whenever any
constituent was active.

The screening null sampled proteins together with their complete test blocks to preserve
within-protein dependence. Across 468 tested pathways and 10,000 permutations, nine pathways
passed the BH and permutation criteria ($q=0.008$--0.026). Several involved cytochrome P450
and drug metabolism. Five drug-metabolism pathways remained significant across coverage
thresholds of 10, 20, and 50 tested constituents. The released interpretation is a
screening-derived interaction hypothesis requiring exposure and biological-context evaluation;
it is not an observed clinical interaction or a therapeutic-efficacy claim.

\begin{table}[htbp]
\centering\small
\caption{Case-study contrasts. Pathway counts use their stated test families. Disease-overlap
rows use the whole Reactome background and Open Targets 26.09, with channel score $\geq0.5$.}
\label{tab:case}
\begin{tabularx}{\textwidth}{P{0.27\textwidth}P{0.29\textwidth}Y}
\toprule
\rowcolor{impflight}
\textbf{Analysis specification} & \textbf{Result} & \textbf{Permitted interpretation}\\
\midrule
Potency hits; whole annotation background & 12,155 background proteins; 66/1,684 pathways
significant & Background-sensitive enrichment requiring assay-coverage control.\\\addlinespace
Potency hits; assayed-protein background & 391 background proteins; 0/959 pathways significant &
No pathway passes the prespecified procedure; no positive mechanism claim released.\\\addlinespace
Active/inactive screening; $\geq20$ constituents per protein & 472 proteins; 51,304 tests;
9/468 pathways significant & Screening-derived hypotheses, including drug-metabolism signals.\\\addlinespace
T2DM genetic-association channel & Overlap 9; fold 0.97; $p=0.59$ & No detected enrichment
under this disease-evidence definition.\\\addlinespace
T2DM literature co-mention channel & Overlap 57; fold 5.26; $p=5.2\times10^{-26}$ &
Literature-defined overlap with potential shared-source and study-selection dependence.\\
\bottomrule
\end{tabularx}
\end{table}

\subsection{Disease evidence is a controlled analytic variable}
For type 2 diabetes mellitus (T2DM; \code{MONDO:0005148}), Open Targets 26.09 supplied
separate genetic-association and literature co-mention sets. At a channel score threshold
of $0.5$, the genetic set contained 478 proteins within the 12,155-protein Reactome background;
nine overlapped the 237 potency-qualified targets (fold enrichment 0.97; hypergeometric
$p=0.59$). The literature set contained 556 background proteins and overlapped 57 targets
(fold enrichment 5.26; $p=5.2\times10^{-26}$).

These are different evidence-defined questions. Strong literature overlap can reflect
repeated study of familiar proteins and dependence between text-mined associations and
compound--target literature. It does not substitute for independent genetic support.
The background must also remain attached to the statistic: with the 391-protein assayed
background, 14 proteins met the genetic criterion, the overlap remained nine, and the
corresponding result was fold 1.06 and $p=0.50$. Reporting the background explicitly prevents
these distinct calculations from being mistaken for inconsistent estimates of one quantity.

The case demonstrates three architectural functions: source typing determines what a
target--disease edge means; the compiled analysis records which population defines the
statistical comparison; and release preserves the distinction between an observed
computational signal and a biological or clinical conclusion. The system can therefore
report an informative negative analytic result and a separately supported screening
hypothesis without merging them into a single mechanism narrative.

\section{Governance Regression and Ablation}
\subsection{Corpus and evaluation design}
We evaluated governance using scientific mutation testing. Ten compliant base outputs
(six Chinese and four English) were paired with structured claims and source records.
Mutation operators introduced one named scientific error at a time, producing
\NMutants{} erroneous outputs across \NErrorClasses{} classes. Errors included population
and endpoint substitution, altered direction or certainty, predictions presented as clinical
findings, fabricated or tampered citations, near-name medicinal substitution, dose errors,
formula--constituent transfer, processing-state transfer, unsupported human exposure,
duplicate-study counting, and unmeasured outcomes presented as negative findings.

The same artifacts were evaluated under no governance, domain-only governance, kernel-only
governance, and the full stack. The kernel contributed evidence-ingest, output, and
claim-kind licensing gates; the domain layer contributed the TCM claim contract and
snapshot-path release check. Each gate was evaluated once per case, and configurations
were assembled from the enabled gate verdicts while preserving the ingest-to-output
dependency. This paired design isolates which checks change release decisions; it does not
measure model generation quality or end-to-end discovery success.

\subsection{Layer-wise results and finite-sample uncertainty}
Table~\ref{tab:ablation} reports the observed release decisions. Without governance, all
121 erroneous outputs were released. The domain layer released 57 (47.11\%), the kernel
released seven (5.79\%), and the full stack released none. All ten compliant outputs were
retained in every configuration. Thus, the combined system intercepted the full constructed
error corpus, while neither layer alone did so.

For consistent uncertainty reporting, all intervals in Table~\ref{tab:ablation} are
two-sided 95\% Clopper--Pearson binomial intervals recalculated from the displayed counts.
For zero observed events in $n$ trials, the upper bound is
\begin{equation}
u_{0,n}=1-(0.025)^{1/n}.
\label{eq:ci}
\end{equation}
Equation~\ref{eq:ci} gives 3.00\% for $0/121$ erroneous releases and 30.85\% for $0/10$ false refusals.
The ten compliant controls therefore provide only a small check on over-refusal. The
intervals are descriptive binomial reference intervals: mutants share base cases and were
constructed by the gate developers, so they do not estimate a population error rate for
independent real-world tasks. The corpus also informed gate repair, making this a regression
evaluation rather than a held-out generalization test.

\begin{table}[htbp]
\centering\small
\caption{Layer-wise ablation on the same 121 mutants and ten compliant controls. Intervals
use the two-sided exact binomial method. Counts are existing regression results; intervals
are recomputed uniformly for this report.}
\label{tab:ablation}
\begin{tabularx}{\textwidth}{Yrrr}
\toprule
\rowcolor{impflight}
\textbf{Configuration} & \textbf{Errors released} & \textbf{95\% CI (\%)} & \textbf{False refusals}\\
\midrule
No governance & 121/121 (100.00\%) & 97.00--100.00 & 0/10\\
Domain layer only & 57/121 (47.11\%) & 37.97--56.39 & 0/10\\
Kernel only & 7/121 (5.79\%) & 2.36--11.56 & 0/10\\
\textbf{Full stack} & \textbf{0/121 (0.00\%)} & \textbf{0.00--3.00} & \textbf{0/10}\\
\bottomrule
\end{tabularx}
\par\vspace{4pt}\parbox{\textwidth}{\footnotesize The false-refusal rate is 0.00\% in all
four configurations, with a 95\% exact interval of 0.00--30.85\% for each configuration.
The four rows reuse the same controls and do not constitute 40 independent observations.}
\end{table}

\subsection{Gate-wise attribution}
The leave-one-gate-out results in Table~\ref{tab:gates} identify complementary mechanisms.
Removing the output gate released 42 errors, including textual drift from structured claims,
direction changes, dose errors, and unsupported citations. Removing evidence-ingest
verification released ten tampered-record cases. Removing the domain claim contract
released six errors involving evidence independence, medicinal identity, exposure, or
unmeasured outcomes. Removing the snapshot-path release check released one unsupported
direction claim.

\begin{table}[htbp]
\centering\small
\caption{Leave-one-gate-out attribution relative to the full stack. All remaining gates stay
enabled. Counts are additional erroneous releases among the same 121 mutants and must not
be summed as disjoint error classes.}
\label{tab:gates}
\begin{tabularx}{\textwidth}{P{0.25\textwidth}rY}
\toprule
\rowcolor{impflight}
\textbf{Gate removed} & \textbf{Additional releases} & \textbf{Failure exposed}\\
\midrule
Output validation & 42 & Wording--claim drift, altered direction or dose, and unsupported
citations or quotations.\\\addlinespace
Evidence-ingest verification & 10 & Records modified after signing.\\\addlinespace
Domain claim contract & 6 & Duplicate evidence, medicinal identity, exposure transfer,
and an unmeasured outcome treated as absent.\\\addlinespace
Snapshot-path release check & 1 & A claimed direction unsupported by the cited relation path.\\\addlinespace
Claim-kind licensing & 0 & Its tested errors were also rejected by remaining gates.\\
\bottomrule
\end{tabularx}
\end{table}

The zero incremental count for licensing indicates overlap on this corpus, not evidence
that the licensing check is unnecessary. Compilation rejects invalid claim types before
execution, while the output and domain checks catch overlapping violations at later
boundaries. Together, the layer and gate ablations show why provenance checking, textual
alignment, and TCM-specific semantics belong to a common release path.

\section{Discussion}
TCMScience makes scientific interpretation an executable obligation. The model remains
responsible for proposing useful analyses and explanations, while source state, operation
authority, and claim support are evaluated by explicit runtime components. This separation
reduces dependence on a model's willingness to follow a cautionary prompt and makes a
refusal traceable to a specific violated contract.

The case study and regression corpus support different conclusions. The case study shows
how background choice, inactive records, and disease-evidence channels alter the permissible
interpretation of one formula analysis. The ablation shows complementary interception of
known constructed failure modes. Neither result establishes universal hallucination
prevention, clinical effectiveness, or superiority on general biomedical task benchmarks.
Independent research-task evaluation is required to measure those broader properties.

The formal specification in Section~4 clarifies the intended distinction between supporting,
contextual, and opposing evidence and between conjunctive dependencies and alternative
support. The reported regression counts evaluate the implemented gate configurations on
their fixed corpus; they are not additional validation of every evidence-composition case
in the specification. Source identity, study quality, claim scope, and statistical validity
remain separate obligations, so a source hash or a high model confidence cannot substitute
for evidence that supports the conclusion.

The principal architectural advantage is therefore compositional control: the same
authority ceiling governs planning and execution, and the same evidence contract governs
structured claims and released wording. This permits new models, tools, and datasets to
extend the scientific workspace while preserving an inspectable boundary around the
conclusions they can justify.

\section{Conclusion}
TCMScience is an autonomous scientific knowledge-discovery agent dedicated to traditional
Chinese medicine. Its trusted kernel, scientific compiler, and claim-support contracts
connect heterogeneous research resources to a governed path from hypothesis to executable
analysis and evidence-bounded conclusion. The network-pharmacology case demonstrates that
source semantics and statistical controls change scientific interpretation, while the
regression ablation establishes complementary error interception by the kernel and TCM
domain layer on the evaluated corpus. By separating generative reasoning from execution
and release authority, TCMScience provides a reproducible technical foundation for
scientific discovery across TCM and biomedicine.

\section*{Code and Data Availability}
TCMScience Studio is available at \url{https://science.impf.ai}. Source code, including
PSH-Harness, BioScience-Harness, and Studio, is available at
\url{https://github.com/IMPF-AI-Institute/TCMScience}. The regression counts are provided in
\path{BioScience-Harness/benchmarks/ablation/results.json}; snapshot construction and
network-pharmacology execution are implemented in
\path{BioScience-Harness/scripts/build_source_snapshots.py} and
\path{BioScience-Harness/scripts/run_network_pharmacology.py}. Upstream biomedical data
retain their source-specific access and reuse conditions. The LaTeX package includes the
regression counts used in the tables and the calculation of their exact binomial intervals.

\section*{Ethics Statement}
This report describes computational architecture, constructed regression cases, and secondary
analysis of public biomedical database records. It reports no recruitment of human
participants, collection of identifiable patient data, or animal experimentation.

\FloatBarrier
\Needspace{10\baselineskip}
{\raggedright
\bibliographystyle{unsrt}
\bibliography{references,supplementary-references}
}
\clearpage
\appendix
\numberwithin{equation}{section}
\renewcommand{\thetable}{\Alph{section}\arabic{table}}
\input{appendices}
\end{document}

%% file: metrics.tex
\newcommand{\NConnectors}{117}
\newcommand{\NOperations}{444}
\newcommand{\NConnectorDomains}{29}
\newcommand{\NTools}{150}
\newcommand{\NToolDomains}{12}
\newcommand{\NDatasets}{76}
\newcommand{\NDatabases}{136}
\newcommand{\NFormulaRecords}{84,294}
\newcommand{\NSnapshotCards}{8}
\newcommand{\NCoreSkills}{4}
\newcommand{\NMutants}{121}
\newcommand{\NErrorClasses}{26}
\newcommand{\NControls}{10}

%% file: appendices.tex
\section{Notation}
\label{app:notation}
\setcounter{table}{0}
This appendix collects the notation in Equations~\ref{eq:authority}--\ref{eq:ci}
and introduces the statistical symbols used in the derivations below.
Table~\ref{tab:notation} separates Boolean decisions from counts and probabilities.
The policy symbol $\pi$ is not a probability, and an inference path $p$ is distinct
from a statistical $p$-value.

\begingroup\small
\setlength{\tabcolsep}{5pt}
\renewcommand{\arraystretch}{1.10}
\begin{longtable}{P{0.28\textwidth}P{\dimexpr0.72\textwidth-4\tabcolsep\relax}}
\caption{Notation for execution, claim support, and statistical calculations.}
\label{tab:notation}\\
\toprule
\rowcolor{impflight}\textbf{Symbol} & \textbf{Meaning and domain}\\
\midrule
\endfirsthead
\caption[]{Notation (continued).}\\
\toprule
\rowcolor{impflight}\textbf{Symbol} & \textbf{Meaning and domain}\\
\midrule
\endhead
\bottomrule
\endfoot
$\Ind[\cdot]$ & Indicator: one when the stated condition holds, zero otherwise.\\\addlinespace
$o$; $A_{\rm eff}$ & Requested operation; intersection of the host, project, run, and skill action sets.\\\addlinespace
$c$, $r$, $\pi$ & Candidate claim, execution run, and governing policy.\\\addlinespace
$e$; $C(c)$ & Evidence record; finite set of all records cited for $c$, with supporting, contextual, or opposing roles.\\\addlinespace
$\mathcal{S}_{\pi}(c)$; $S$ & Family of admissible candidate supporting sets; a nonempty member containing only records offered as support.\\\addlinespace
$V(e)$ & Boolean verification of record identity, integrity, and source binding.\\\addlinespace
$\mathrm{Suff}_{\pi}(S,c)$ & Boolean satisfaction of the complete support contract by $S$ for claim $c$.\\\addlinespace
$G_{\pi}(r,c,C)$ & Boolean execution, artifact, scope, and whole-citation-set checks.\\\addlinespace
$\mathrm{Release}_{\pi}(c,r)$ & Final Boolean release decision in Equation~\ref{eq:release}.\\\addlinespace
$p$, $a$; $\mathcal{P}_{\pi}(c)$ & Supporting derivation, indispensable dependency within it, and family of candidate derivations.\\\addlinespace
$\mathrm{Admissible}_{\pi}(a)$ & Boolean validity of a dependency, including its allowed evidence-to-claim transition.\\\addlinespace
$\mathrm{PathValid}_{\pi}(p)$;\newline $\mathrm{Sufficient}_{\pi}(p,c)$ & Validity of all indispensable dependencies; adequacy of the complete derivation for the requested claim.\\\addlinespace
$\mathrm{Supported}_{\pi}(c)$ & Existence of a complete valid and sufficient derivation in Equation~\ref{eq:paths}.\\\addlinespace
$U$, $H$, $D$ & Declared protein universe, hit set within $U$, and pathway or disease-associated set within $U$.\\\addlinespace
$N$, $n_H$, $K$, $k$ & $|U|$, $|H|$, $|D|$, and observed overlap $|H\cap D|$, respectively.\\\addlinespace
$X$; $p_{\rm HG}$; $F$ & Random overlap under the hypergeometric null, its upper-tail probability, and observed fold enrichment.\\\addlinespace
$m$, $p_{(i)}$, $q_{(i)}$ & Number of tested hypotheses, ordered raw $p$-value, and corresponding BH-adjusted value.\\\addlinespace
$\tau$, $R$ & Declared multiple-testing threshold and number of random permutations.\\\addlinespace
$a_j$, $t_j$ & Active and total classified compound--protein tests for protein $j$; $0\leq a_j\leq t_j$.\\\addlinespace
$x$, $n$, $\theta$, $\alpha$ & Event count, trial count, binomial event probability, and interval tail-error total; here $\alpha=0.05$.\\\addlinespace
$Q_{u}(a,b)$; $L_{x,n}$, $U_{x,n}$ & $u$-quantile of a Beta$(a,b)$ distribution; lower and upper exact interval endpoints.\\
\end{longtable}
\endgroup

\clearpage
\section{Derivations for Authority and Claim Release}
\label{app:release}
\setcounter{table}{0}
The following properties follow from the formal specification in the main text.
They concern the stated set and Boolean semantics; they do not assert additional
implementation coverage or introduce new empirical validation.

\subsection{Authority contraction under additional restrictions}
Index the four authority scopes by $i\in I=\{\mathrm{host},\mathrm{project},
\mathrm{run},\mathrm{skill}\}$. Suppose a revised envelope satisfies
$A'_i\subseteq A_i$ at every scope. Then
\begin{equation}
A'_{\rm eff}=\bigcap_{i\in I}A'_i
\subseteq\bigcap_{i\in I}A_i=A_{\rm eff}.
\label{eq:app-authority}
\end{equation}
Indeed, an operation in every $A'_i$ is in every corresponding $A_i$.
Consequently, $\Ind[o\in A'_{\rm eff}]\leq\Ind[o\in A_{\rm eff}]$.
For a child run, an inherited restriction $A_{\rm child}\subseteq A_{\rm parent}$
preserves this inclusion at each delegation step. Induction over the delegation
chain therefore bounds every descendant by the original envelope. This result
requires actual enforcement of the inherited sets; it is not an operating-system
isolation guarantee.

\subsection{Factorization of the release condition}
For a fixed claim and run, abbreviate the three obligations as
\begin{equation}
\begin{aligned}
E(c)&=\Ind[C(c)\ne\varnothing]\prod_{e\in C(c)}V(e),\\
W_{\pi}(c)&=\Ind\!\left[\exists S\in\mathcal{S}_{\pi}(c):
\mathrm{Suff}_{\pi}(S,c)=1\right],\\
J_{\pi}(c,r)&=G_{\pi}(r,c,C(c)).
\end{aligned}
\label{eq:app-factors}
\end{equation}
Equation~\ref{eq:release} becomes
\begin{equation}
\mathrm{Release}_{\pi}(c,r)=E(c)W_{\pi}(c)J_{\pi}(c,r).
\label{eq:app-release-product}
\end{equation}
Since each factor belongs to $\{0,1\}$, release equals one if and only if all
three factors equal one. In particular,
\begin{equation}
\mathrm{Release}_{\pi}(c,r)\leq E(c),\qquad
\mathrm{Release}_{\pi}(c,r)\leq W_{\pi}(c),\qquad
\mathrm{Release}_{\pi}(c,r)\leq J_{\pi}(c,r).
\label{eq:app-necessary}
\end{equation}
Verification alone is insufficient: $E(c)=1$ can coexist with $W_{\pi}(c)=0$.
A sufficient supporting set is also insufficient if a required whole-set check
fails, for example because an unresolved opposing result affects the claim.
When $C(c)=\varnothing$, $E(c)=0$ despite the empty product being one.
When $\mathcal{S}_{\pi}(c)=\varnothing$, $W_{\pi}(c)=0$ because the existential
statement has no witness.

\subsection{Serial obligations and alternative derivations}
For a finite candidate family, Boolean conjunction and disjunction can be written
using products. Define
\begin{equation}
\begin{aligned}
z_p&=\mathrm{Sufficient}_{\pi}(p,c)
       \prod_{a\in p}\mathrm{Admissible}_{\pi}(a),\\
\mathrm{Supported}_{\pi}(c)&=1-\prod_{p\in\mathcal{P}_{\pi}(c)}(1-z_p).
\end{aligned}
\label{eq:app-path-product}
\end{equation}
The second line equals zero precisely when every $z_p$ is zero; it equals one
when any complete derivation succeeds. An indispensable invalid dependency sets
its own branch to zero. It does not set an independent sufficient alternative
to zero. For example, if a declared inference requires both $e_1$ and $e_2$,
while $e_3$ can independently support the same scoped conclusion, the support
structure is $(v_1\wedge v_2)\vee v_3$, where $v_i$ denotes the applicable
admissibility and support obligations, not merely source verification.

A derivation with no dependencies cannot obtain support from an empty product
alone: it must still satisfy $\mathrm{Sufficient}_{\pi}(p,c)$ and the nonempty
support requirement. Equation~\ref{eq:app-path-product} evaluates to zero when
there are no candidate derivations. For set-based and path-based descriptions
to agree, each successful supporting set must have a complete admissible
inference witness, and each successful witness must bind a permitted nonempty
supporting set. This correspondence is a requirement on their representation,
not a consequence of citation counts.

\subsection{Context, opposition, and changes to the citation set}
Consider a verified clinical study used as support and a verified docking record
used as context. If the clinical study satisfies the scoped support contract,
the contextual record need not become an indispensable clinical dependency.
Both records still enter $E(c)$ and the role-sensitive checks in $J_{\pi}(c,r)$.
Changing the claim to a causal assertion about the predicted interaction changes
the support obligations and requires revalidation.

Holding $E$ and $J$ fixed, adding an independently sufficient branch cannot make
$W_{\pi}$ decrease. Release, however, is not generally monotone in the number
of citations: adding an unverified record makes $E=0$, and adding an unresolved
contradictory result can make $J=0$. Repeated database records of one study do
not establish independence. Identity deduplication and the synthesis contract
must determine which records can supply distinct evidential contributions.

\section{Statistical Derivations for the Case Study}
\label{app:statistics}
\setcounter{table}{0}
This section makes explicit the calculations underlying the reported overlap
comparisons and the distinction between background control, multiple testing,
and block-preserving screening analyses. Numerical examples reuse the counts
in Section~\ref{sec:case}.

\subsection{Hypergeometric overlap and fold enrichment}
Let the fixed background $U$ contain $N$ proteins, with a set $D$ of $K$ proteins
carrying the annotation or disease-evidence property of interest. Under a null
that selects $n_H$ proteins uniformly without replacement, there are
$\binom{N}{n_H}$ possible hit sets. Exactly $x$ overlaps with $D$ can be formed
by choosing $x$ proteins from $D$ and $n_H-x$ from its complement. Thus
\begin{equation}
\Pr(X=x)=\frac{\binom{K}{x}\binom{N-K}{n_H-x}}{\binom{N}{n_H}},
\quad \max(0,n_H+K-N)\leq x\leq\min(n_H,K).
\label{eq:app-hypergeom}
\end{equation}
The one-sided overrepresentation probability for observed overlap $k$ and the
fold enrichment are
\begin{equation}
\begin{aligned}
p_{\rm HG}&=\Pr(X\geq k)
 =\sum_{x=k}^{\min(n_H,K)}
   \frac{\binom{K}{x}\binom{N-K}{n_H-x}}{\binom{N}{n_H}},\\
\mathbb{E}[X]&=\frac{n_H K}{N},\qquad
F=\frac{k}{\mathbb{E}[X]}=\frac{k/n_H}{K/N}.
\end{aligned}
\label{eq:app-enrichment}
\end{equation}
The expectation follows by summing inclusion indicators, each with marginal
probability $n_H/N$, over the $K$ annotated proteins. The fold is defined when
$n_HK>0$. Degenerate sets must be handled explicitly rather than assigned a
finite fold by division by zero.

Table~\ref{tab:overlap-check} reproduces the disease-channel comparisons from
the reported integers. The genetic comparison changes from
$9/(237\times478/12{,}155)=0.97$ to $9/(237\times14/391)=1.06$
when the universe changes. The same overlap count therefore need not imply
the same enrichment.

\begin{table}[htbp]
\centering\small
\caption{Arithmetic reconstruction of the reported disease-overlap comparisons.
$p_{\rm HG}$ is an unadjusted upper-tail probability.}
\label{tab:overlap-check}
\begin{tabularx}{\textwidth}{Yrrrrrr}
\toprule
\rowcolor{impflight}
\textbf{Channel / background} & $N$ & $K$ & $n_H$ & $k$ & $F$ & $p_{\rm HG}$\\
\midrule
Genetic / Reactome & 12,155 & 478 & 237 & 9 & 0.97 & 0.59\\
Literature / Reactome & 12,155 & 556 & 237 & 57 & 5.26 & $5.2\times10^{-26}$\\
Genetic / assayed & 391 & 14 & 237 & 9 & 1.06 & 0.50\\
\bottomrule
\end{tabularx}
\end{table}

Changing the background requires recomputing $H\cap U$, $D\cap U$, and all four
counts. In the pathway analysis, the original annotation-size filter of
5--500 proteins is applied before restriction to the declared universe.
The tested families of 1,684 and 959 pathways therefore cannot be treated as
one unchanged family with only a different denominator.

\subsection{Benjamini--Hochberg adjustment}
For a declared family of $m$ tests, order the raw probabilities as
$p_{(1)}\leq\cdots\leq p_{(m)}$. The BH adjusted values are
\begin{equation}
q_{(i)}=\min\!\left\{1,\min_{j=i,\ldots,m}\frac{m}{j}p_{(j)}\right\}.
\label{eq:app-bh}
\end{equation}
At threshold $\tau$, the step-up procedure finds
$i^*=\max\{i:p_{(i)}\leq i\tau/m\}$ and rejects the first $i^*$ hypotheses;
if the set is empty, it rejects none. The reverse cumulative minimum in
Equation~\ref{eq:app-bh} gives the same decisions through $q_{(i)}\leq\tau$.
The test family must be specified before selecting discoveries. Removing
nonsignificant pathways before adjustment would change the procedure.
The original BH guarantee assumes independent test statistics\cite{bh};
overlapping pathways and assay selection require explicit attention to the
null model and dependence structure. Adjustment alone does not remove those
sources of bias.

\subsection{Why active and inactive tests must remain together}
For protein $j$, let $t_j$ be its number of classified tests and $a_j$ the active
count. For a selected protein set $P$, its test-level active fraction is
\begin{equation}
\widehat f(P)=\frac{\sum_{j\in P}a_j}{\sum_{j\in P}t_j}.
\label{eq:app-active-fraction}
\end{equation}
This descriptive fraction differs from both the unweighted average of
$a_j/t_j$ and the fraction of proteins with at least one active result.
At the coverage threshold of 20 constituents, the reported totals give
$3{,}357/51{,}304=6.54\%$. Collapsing each protein to $\Ind[a_j>0]$ would discard
inactive outcomes and the unequal numbers of tested constituents.

The reported screening null samples proteins with their complete test blocks.
In a general block-permutation formulation, a declared statistic $T$ is computed
for the observed pathway and for $R$ admissible random draws. Blocks preserve
within-protein dependence; eligibility, pathway size, and any required matching
constraints belong to the sampling rule. For a random-draw upper-tail test that
includes ties, a standard finite-simulation estimate is\cite{phipson2010}
\begin{equation}
\widehat p_{\rm perm}=
\frac{1+\sum_{b=1}^{R}\Ind[T^{(b)}\geq T^{\rm obs}]}{R+1}.
\label{eq:app-permutation}
\end{equation}
For $R=10{,}000$, this convention has minimum probability $1/10{,}001$.
Equation~\ref{eq:app-permutation} states a reproducible random-permutation
convention, not a reconstruction of unprovided draws. Recovering the reported
nine discoveries additionally requires the original statistic, admissible
sampling rule, random seed, pathway membership, and joint BH/permutation
criterion. Aggregate counts alone do not determine those results.

\clearpage
\section{Exact Binomial Intervals and Their Interpretation}
\label{app:intervals}
\setcounter{table}{0}
Under a binomial reference model, let $X\sim\mathrm{Binomial}(n,\theta)$ and
observe $X=x$. A two-sided Clopper--Pearson interval\cite{clopper1934} inverts the binomial tails:
for interior counts, its endpoints satisfy
\begin{equation}
\Pr_{\theta=L_{x,n}}(X\geq x)=\frac{\alpha}{2},\qquad
\Pr_{\theta=U_{x,n}}(X\leq x)=\frac{\alpha}{2}.
\label{eq:app-tail-inversion}
\end{equation}
Using the binomial-tail identities with Beta distributions yields
\begin{equation}
\begin{aligned}
L_{x,n}&=\begin{cases}
0,&x=0,\\
Q_{\alpha/2}(x,n-x+1),&x>0,
\end{cases}\\[3pt]
U_{x,n}&=\begin{cases}
1,&x=n,\\
Q_{1-\alpha/2}(x+1,n-x),&x<n.
\end{cases}
\end{aligned}
\label{eq:app-cp}
\end{equation}
At zero observed events, $\Pr_{\theta}(X=0)=(1-\theta)^n$.
Solving $(1-U_{0,n})^n=\alpha/2$ gives the main-text expression
\begin{equation}
U_{0,n}=1-\left(\frac{\alpha}{2}\right)^{1/n}.
\label{eq:app-zero-events}
\end{equation}
For $x=n$, the symmetric boundary is $L_{n,n}=(\alpha/2)^{1/n}$.
At $\alpha=0.05$, the zero-event upper bounds are $3.0027\%$ for $n=121$
and $30.8497\%$ for $n=10$, which round to the displayed $3.00\%$ and $30.85\%$.
These are two-sided 95\% endpoints; replacing $\alpha/2$ by $\alpha$ would
instead compute a one-sided 95\% upper bound.

Table~\ref{tab:interval-check} applies the same formula to every reported
configuration. The event is an erroneous release for mutant cases and a false
refusal for compliant controls.
\begin{table}[htbp]
\centering\small
\caption{Exact interval calculations from the existing regression counts.
All endpoints are percentages.}
\label{tab:interval-check}
\begin{tabularx}{\textwidth}{Yrr}
\toprule
\rowcolor{impflight}
\textbf{Configuration / event} & \textbf{Count} & \textbf{Two-sided 95\% interval}\\
\midrule
No governance / erroneous release & 121/121 & 97.00--100.00\\
Domain layer / erroneous release & 57/121 & 37.97--56.39\\
Kernel / erroneous release & 7/121 & 2.36--11.56\\
Full stack / erroneous release & 0/121 & 0.00--3.00\\
Any configuration / false refusal & 0/10 & 0.00--30.85\\
\bottomrule
\end{tabularx}
\end{table}

The package script \path{recompute_intervals.py} evaluates the Beta quantiles
from \path{data/governance_counts.json}. The added
\path{check_appendix_calculations.py} also checks the overlap arithmetic and
compares the interval endpoints with a binomial-tail implementation.
Neither script runs a new benchmark or enlarges the evaluation corpus.

Exactness refers to coverage under the binomial model. The constructed mutants
share base cases and informed gate repair; their descriptive reference intervals
do not establish an independent real-world failure rate. The four configurations
reuse the same ten controls, so pooling them as $n=40$ would overstate the
information about false refusals. Likewise, a zero incremental count in a
gate-removal analysis measures redundancy on this corpus, rather than proving
that the gate has no distinct purpose.